\documentclass[preprint,12pt]{elsarticle}
\usepackage{xcolor}%
\usepackage{amssymb}
\usepackage{amsmath}
\usepackage[flushleft]{threeparttable}
\usepackage{subcaption}
\usepackage{booktabs}
\usepackage{longtable}
\usepackage{adjustbox}

\newcommand{\jiaxi}[1]{{\color{blue}[Jiaxi: #1]}}
\newcommand{\xl}[1]{{\color{green}[Xiaoxiao: #1]}}

\newcommand{\eg}{{\em e.g.,~}}           

\newcommand{\etal}{{\em et al.~}}

\journal{xxx}

\begin{document}

\begin{frontmatter}

\title{Towards Trustworthy Foundation Models for Medical Image Analysis}

\author{
Congzhen~Shi$^{a,d,\dagger}$,
Ryan~Rezai$^{b,\dagger}$,
Jiaxi~Yang$^{a,\dagger}$,
Qi~Dou$^{c}$,
Xiaoxiao Li$^{a,d,*}$
}

\affiliation{organization={The University of British Columbia}}
\affiliation{organization={University of Waterloo}}
\affiliation{organization={The Chinese University of Hong Kong}}
\affiliation{organization={Vector Institute}}

\begin{abstract}
The rapid advancement of foundation models in medical imaging represents a significant leap toward enhancing diagnostic accuracy and personalized treatment. However, the deployment of foundation models in healthcare necessitates a rigorous examination of their trustworthiness, encompassing privacy, robustness, reliability, explainability, and fairness. The current body of survey literature on foundation models in medical imaging reveals considerable gaps, particularly in the area of trustworthiness. 
Additionally, existing surveys on the trustworthiness of foundation models do not adequately address their specific variations and applications within the medical imaging domain. This survey aims to fill that gap by presenting a novel taxonomy of foundation models used in medical imaging and analyzing the key motivations for ensuring their trustworthiness. We review current research on foundation models in major medical imaging applications, focusing on segmentation, medical report generation, medical question and answering (Q\&A), and disease diagnosis. These areas are highlighted because they have seen a relatively mature and substantial number of foundation models compared to other applications. We focus on literature that discusses trustworthiness in medical image analysis manuscripts. We explore the complex challenges of building trustworthy foundation models for each application, summarizing current concerns and strategies for enhancing trustworthiness. Furthermore, we examine the potential of these models to revolutionize patient care. 
Our analysis underscores the imperative for advancing towards trustworthy AI in medical image analysis, advocating for a balanced approach that fosters innovation while ensuring ethical and equitable healthcare delivery.
\end{abstract}


\end{frontmatter}

\section{Introduction}
With the advancement of foundational models, the field of medical image analysis is poised at the brink of a revolution. Foundation models are large-scale machine learning models trained on extensive and diverse datasets. Following their initial training, these models can be adapted to specific downstream tasks with minimal adjustments. By leveraging their extensive pre-training on large-scale datasets, they offer unprecedented analytical depth, enabling the analysis and prediction of medical images ranging from radiology to pathology. The integration of foundation models into medical image analysis holds the potential to enhance diagnostic accuracy, expedite treatment schedules, and ultimately enhance patient outcomes.

The integration of foundation models into medical image analysis garnered significant interest, leading to a surge of impactful studies in the field. Notably, several perspective papers have emerged, highlighting the potential and future directions of leveraging foundation models in the medical domain. Research in foundation models for medical imaging has been explored in many areas, for example, precise segmentation and detection of tumors \cite{ma2024segment}, auto-organ segmentation \cite{zhang2023segment}, generating clinical reports \cite{thawkar2023xraygpt}, extracting quantitative features from medical images using deep learning to predict disease characteristics and outcomes \cite{zhao2023clip}, and medical Q\&A systems~\cite{chen2024chexagent}.

\begin{figure*}[t]
    \centering
    \includegraphics[width=\linewidth]{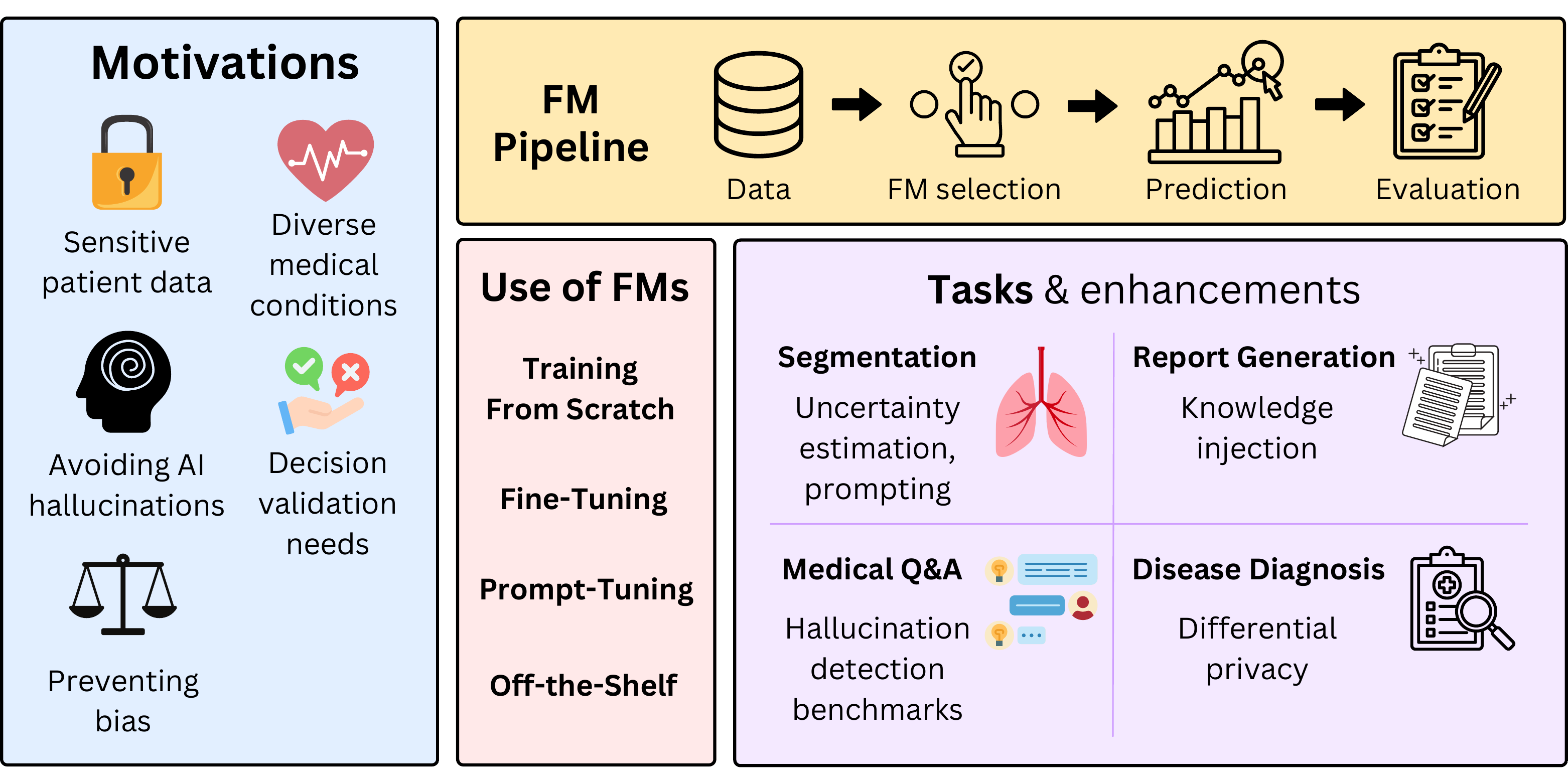}
    \caption{A schematic overview of motivations, foundation model usage, tasks and trustworthiness enhancements discussed in this paper.}
    \label{fig:overview}
\end{figure*}

However, the deployment of foundation models in such a critical sector raises significant concerns regarding their trustworthiness, which encompasses privacy, robustness, reliability, explainability, and fairness (detailed definition see Sec~\ref{trustworthiness}). In medical contexts, where decisions have profound implications on patient health, ensuring the trustworthiness of foundation models becomes paramount. It involves rigorous validation against clinical standards, continuous monitoring for performance drift, and mechanisms to interpret model decisions transparently. To fill this gap, we review and discuss recent advances of trustworthiness in foundation models for medical image analysis.  To the best of our knowledge, this is the first survey of foundation models for medical image analysis from the \textit{trustworthiness} perspective, which is different from the existing surveys of foundation models for medical image analysis. The schematic overview in Fig.~\ref{fig:overview} illustrates how foundation models can be integrated into medical image analysis tasks, while also emphasizing the need to address trustworthiness issues in these applications.

\noindent\textbf{Comparison with existing surveys.} We compare existing surveys that discuss similar topics on foundation models, trustworthiness, and particularly those focused on medical imaging. Our survey paper exhibits several distinct advantages over the existing literature, as illustrated in Table~\ref{tab:comparison}. Existing survey papers on foundation models in medical imaging exhibit significant gaps, particularly in addressing trustworthiness issues. For instance, while Azad \etal~\cite{azad2023foundational}, Zhao \etal~\cite{zhao2023clip}, and He \etal~\cite{he2024foundation} discussed foundation models for medical imaging, they fall short in providing detailed analysis on trustworthiness. Whereas, He \etal~\cite{he2024survey}, Sun \etal~\cite{sun2024trustllm}, and Liu \etal~\cite{liu2024trustworthy} survey trustworthiness in foundation models but do not focus on medical imaging applications. Similarly, Salahuddin \etal~\cite{salahuddin2021transparency} and Hasani \etal~\cite{Hasani2022-rm} focus on trustworthiness issues in medical imaging but do not cover the foundation models. Furthermore, most of the existing survey papers fail to provide a comprehensive examination of trustworthiness and detailed insights into medical imaging application-specific challenges and solutions. In contrast, our survey uniquely integrates an in-depth analysis of trustworthiness across both LLMs and Vision foundation models, emphasizing privacy, robustness, reliability, explainability, and fairness, alongside a comprehensive review of their applications in medical imaging. This holistic approach ensures a more detailed and multifaceted understanding of the current landscape and addresses the critical need for trustworthy AI in medical image analysis, offering valuable insights that are not as extensively covered in other surveys.

\begin{table*}
\begin{threeparttable}
\centering
\resizebox{1.0\textwidth}{!}{
\begin{tabular}
{cccccc}
\hline\hline

\textbf{Work} & \textbf{LLMs} & \textbf{Vision foundation models} &
\textbf{Medical imaging} & \multicolumn{2}{c}{\textbf{Trustworthiness}} \\ &&&&\textbf{Application-} & \textbf{Comprehen-} \\
&&&& \textbf{specific analysis$^\dagger$} & \textbf{siveness$^\ddagger$} \\

\hline
Azad~\etal \cite{azad2023foundational} & $\checkmark$ & $\checkmark$ & $\checkmark$ & $\times$ & $\times$\\
\hline
Zhao~\etal\cite{zhao2023clip} & $\times$ & $\checkmark$ & $\checkmark$ & $\times$ & $\times$ \\
\hline
He \etal\cite{he2024foundation} & $\checkmark$ & $\checkmark$ & $\checkmark$ & $\times$ & $\times$\\
\hline
He \etal\cite{he2024survey} & $\checkmark$ & $\times$ & $\times$ & $\times$ & $\checkmark$ \\
\hline
Sun \etal\cite{sun2024trustllm} & $\checkmark$ & $\times$ & $\times$ & $\times$ & $\checkmark$\\
\hline
Liu \etal\cite{liu2024trustworthy} & $\checkmark$ & $\times$ & $\times$ & $\times$ & $\checkmark$\\
\hline
Salahuddin \etal\cite{salahuddin2021transparency} & $\times$ & $\times$ & $\checkmark$ & $\checkmark$ & $\times$ \\
\hline
Hasani \etal\cite{Hasani2022-rm} & $\times$ & $\times$ & $\checkmark$ & $\times$ & $\checkmark$ \\
\hline
\textbf{Ours} & $\checkmark$ & $\checkmark$ & $\checkmark$ & $\checkmark$ & $\checkmark$ \\
\hline
\end{tabular}
}
\caption{Comparison with existing surveys.}
\begin{tablenotes}
\item $\dagger$ the paper illustrates that different applications have different trustworthiness \\ concerns or solutions
\item $\ddagger$ in-depth analysis on more than two different aspects of trustworthiness
\end{tablenotes}
\label{tab:comparison}
\end{threeparttable}
\end{table*}

\noindent\textbf{Contributions.}
The contributions of this paper can be highlighted in the following aspects.

\begin{enumerate}

    \item \textbf{Identifying:} We identify trustworthiness concerns in foundation models for medical image analysis, exploring how these issues manifest across different types of foundation model use and various medical imaging tasks.
    \item \textbf{Surveying}: We conduct an in-depth review of medical image foundation models in the existing literature and categorize them according to the common applications of medical image analysis that use foundation models, including \emph{segmentation, report generation, medical Q\&A,} and \emph{disease diagnosis}, and their use of foundation models.
    \item \textbf{Unveiling}: We uncover and remark on the trustworthiness issues in the existing literature, highlighting prominent trustworthiness concerns associated with each kind of application, and noting the significant gaps in addressing these concerns.
    \item \textbf{Envisioning}: We propose future research directions, emphasizing the need for innovative approaches to enhance model trustworthiness.
\end{enumerate}

\begin{figure*}[t]
    \centering
    \includegraphics[width=1\linewidth]{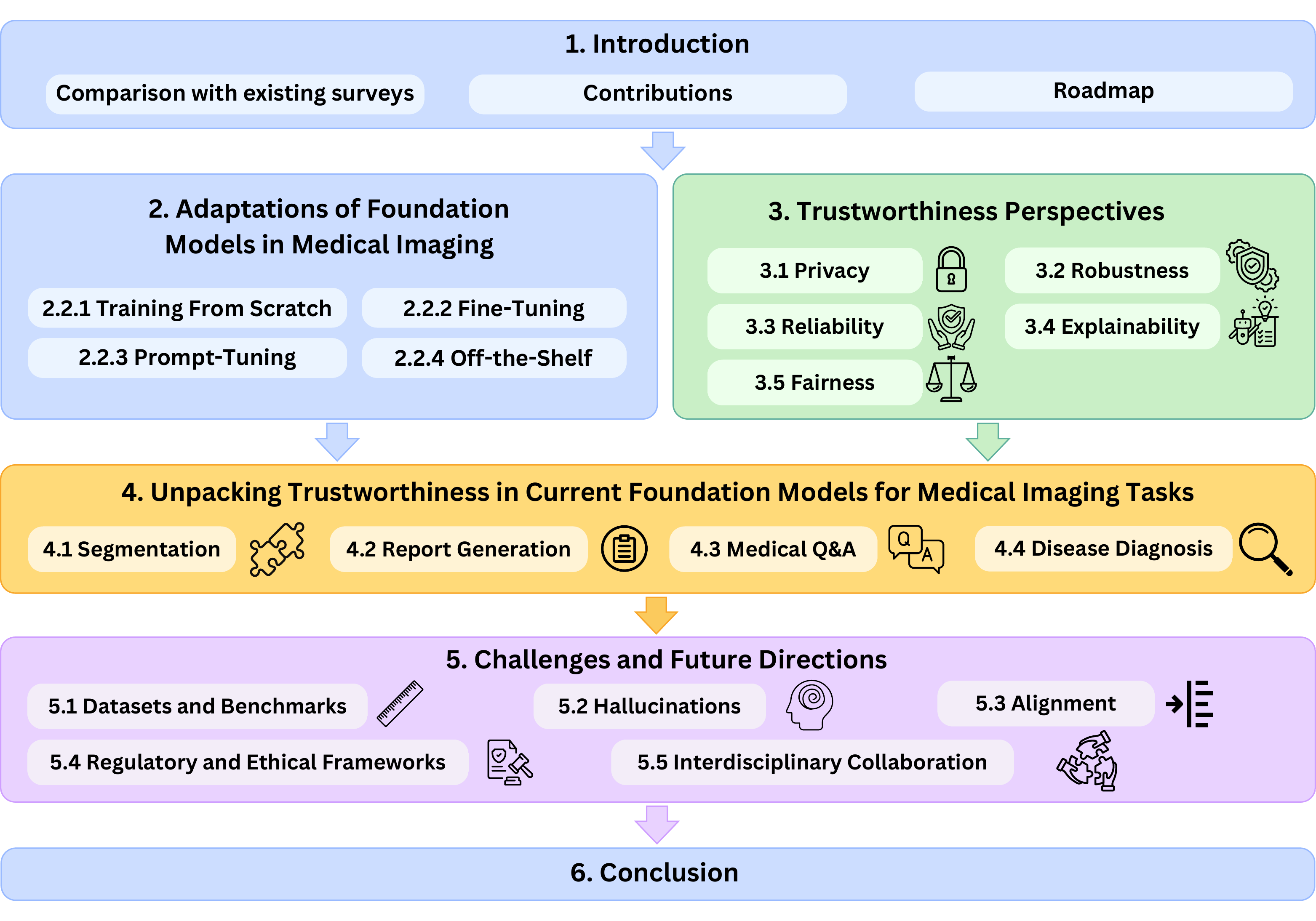}
    \caption{Roadmap of this paper.}
    \label{fig:roadmap}
\end{figure*}

\noindent\textbf{Roadmap.}
The roadmap of this paper is shown in Figure~\ref{fig:roadmap}. The subsequent sections are organized as follows: In Section~\ref{background}, we introduce the background and usage of foundation models for medical image analysis. In Section~\ref{trustworthiness}, we define five categories of trustworthiness issues. In Section~\ref{applications}, we categorize the existing literature of foundation models for medical image analysis based on their application, and then by the reported trustworthiness concerns or strategies for enhancements. Finally, we conclude with challenges and future directions in Section~\ref{challenges} before presenting our conclusion in Section~\ref{conclusion}.

\begin{figure*}[t]
    \centering
    \includegraphics[width=1\linewidth]{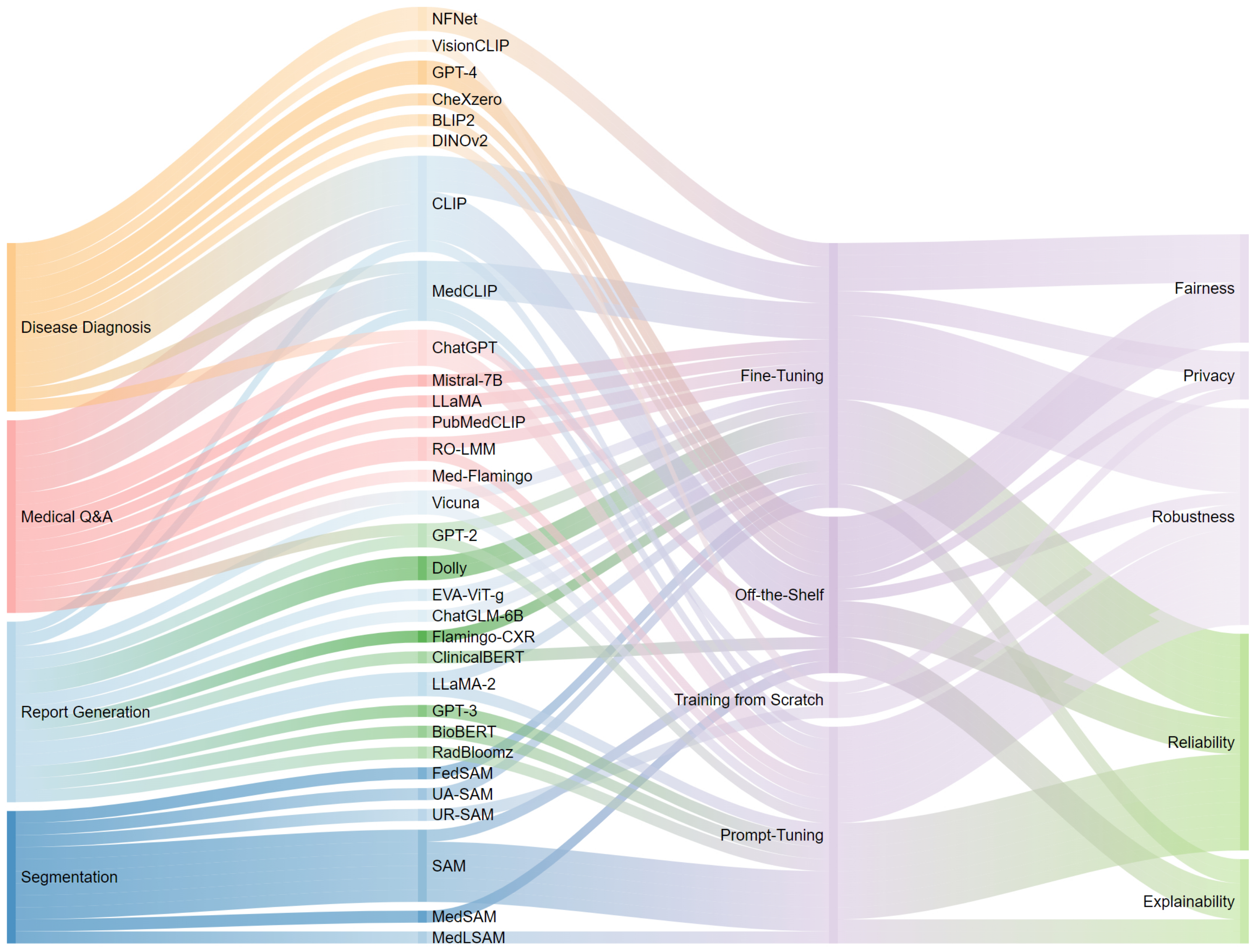}
    \caption{\textbf{Landscape of foundation models, their usage, and trustworthiness issues for medical image analysis.} The relationship between FM usage and trustworthiness issues for different applications is shown. Each thread represents one piece of literature reviewed in this survey.}
    \label{fig:landscape}
\end{figure*}

To provide a comprehensive overview, Figure~\ref{fig:landscape} illustrates the landscape of foundation models for medical image analysis studied in this survey. We have covered 31 recent foundation models with applications in medical imaging from 76 research papers, among which 48 are research papers particularly developing for or adapting to medical imaging, published from 2019 to 2024.

Given that the type of foundation models is intricately linked to specific medical imaging applications, our survey is systematically categorized according to the prevalent tasks in medical imaging. In the four medical imaging areas we focused on—disease diagnosis, medical Q\&A, report generation, and segmentation—each foundation model is associated with multiple medical applications, demonstrating their versatility and applicability across various medical imaging tasks. For instance, models like MedCLIP and ChatGPT are extensively used for disease diagnosis and medical Q\&A, while models like SAM \cite{kirillov2023segment} and its variants are specialized for Segmentation.  LLaMA and Vicuna \cite{vicuna2023} are used in both medical Q\&A and report generation. GPTs  \cite{brown2020language,openai2024gpt4} are used in all three applications except segmentation. CLIP \cite{radford2021learning} and its variants are popular choices for medical Q\&A and disease diagnosis. In this survey, we also investigate the taxonomy of using foundation models, categorized by training from scratch, fine-tuning, prompt-tuning, or direct off-the-shelf use and how they are linked to trustworthiness aspects. The overlapping and intersecting lines between the use of foundation models and trustworthiness issues indicate the multi-faceted challenges and considerations in deploying these models effectively.

\section{Adaptations of Foundation Models in Medical Imaging} 
\label{background}
\subsection{Definition of Foundation Models}
Foundation models refer to massive machine learning models trained on extensive volumes of diverse datasets, which can later be fine-tuned for specific downstream tasks with relatively minimal additional data (\eg ViT \cite{dosovitskiy2021image}). These models (\eg GPT \cite{brown2020language,openai2024gpt4}, SAM \cite{kirillov2023segment}, CLIP \cite{radford2021learning}) serve as versatile platforms that can be adapted to various specific tasks, ranging from language translation and content creation to complex problem-solving in healthcare domains. The adaptability and efficiency of foundation models make them a cornerstone in the development of cutting-edge technologies, providing a robust basis for innovation and research in the field of medical image analysis.

\subsection{Different Methods for Adapting Foundation Models in Medical Imaging}
There are four major approaches for training and using foundation models in medical image analysis: training from scratch, fine-tuning, prompt-tuning, and using off-the-shelf models.

\subsubsection{Training From Scratch}
Foundation models for medical image analysis, often using the transformer architecture, are trained on large, diverse datasets. Some, like the Segment Anything Model (SAM) \cite{kirillov2023segment}, are pre-trained with labeled medical images and paired segmentation masks. Others undergo self-supervised learning (SSL), such as generative SSL (\eg MAE \cite{he2021masked}) and contrastive SSL (\eg DINO \cite{caron2021emerging}, SimCLR \cite{chen2020simple}, CLIP \cite{radford2021learning}),  where they learn generic data patterns without labeled examples. Masked autoencoders (MAEs) \cite{he2021masked} involve pre-text training to learn image representations by treating images as sequences of patches, masking out certain patches, and predicting masked parts of the image, thereby teaching the model to understand visual features such as shapes and textures. Among the contrastive SSL strategies, a popular and advanced approach is contrastive language-image pre-training (CLIP) \cite{radford2021learning}, which uses the pre-training task of predicting which caption goes with which image. After learning a useful generalized representation of visual features, models are fine-tuned on specific downstream tasks, like tumor detection, using a smaller labeled dataset, thereby tailoring their broad capabilities to precise medical diagnosis. 

\subsubsection{Fine-Tuning}
In medical image analysis, fine-tuning foundation models on specific datasets significantly enhances their ability to detect diseases in medical images such as X-rays, MRI scans by adjusting model parameters to the peculiarities of medical images.
Adopting full fine-tuning for medical foundation models involves adjusting all the parameters of a pre-trained model to tailor it to a specific task or dataset. However, this will lead to a computational burden. To address the expensive computation cost, parameter-efficient tuning, such as Lora~\cite{hu2021lora} is proposed to adjust a small set of parameters in a pre-trained model to optimize it for new tasks with minimal computational overhead. This approach is particularly useful in medical image analysis, as most hospitals lack sufficient and powerful computational resources. Another objective for fine-tuning is to align with clinical insights, refine diagnostic accuracy and reliability. As a popular alignment approach, Reinforcement Learning with Human Feedback such as InstructGPT~\cite{ouyang2022training} can be leveraged to use medical experts' feedback for foundation model alignment. Similarly, AI-generated feedback (e.g., Self-Refine~\cite{madaan2023selfrefine}) simulates expert reviews, allowing continuous improvement in interpreting medical images and identifying complex conditions, even without direct human input.

\subsubsection{Prompt-Tuning}
Prompt tuning, or prompt engineering, enhances foundation models for specific tasks by crafting input prompts while keeping the model backbone frozen~\cite{liu2023pre}. The terms ``prompt tuning'' and ``prompt engineering'' are often used interchangeably, although their focuses differ. Prompt tuning focuses on updating continuous embeddings optimized for specific tasks, while prompt engineering involves trying out different prompts to find the ones that work the best. Despite these differences, both approaches are commonly employed together in practice to direct the model's pre-trained knowledge to meet the precise needs of the downstream task, sometimes outperforming full fine-tuning (\eg VPT~\cite{jia2022visual}). In medical image analysis, well-designed prompts are specific inputs, such as text descriptions that guide the medical foundation model to focus on important features or regions within medical images. For example, a well-designed textual prompt for a medical foundation model analyzing chest X-rays might be, ``Focus on the lower lobe of the right lung for potential infiltrates.'' This can significantly improve a model's diagnostic accuracy and provide a resource-efficient way to tailor models for medical diagnostics.

\subsubsection{Off-the-Shelf}
Due to the impressive generalization abilities acquired from the huge amount of training data, many general-purpose foundation models can be applied directly to the medical image analysis domain without any modification. For instance, SAM \cite{kirillov2023segment} can perform remarkably when applied directly to medical image analysis tasks \cite{huang2024segment}. For LLM-based foundation models, off-the-shelf models are generally used to do in-context learning \cite{brown2020language}, which refers to providing demonstrations or examples to the LLM at inference time to enhance its performance.

\section{Trustworthiness Perspectives in Medical Image Analysis}
\label{trustworthiness}

Trustworthiness in foundation models for medical image analysis is essential due to several critical factors and can cover different perspectives. First, \emph{privacy} is paramount, as medical images contain highly sensitive patient information requiring robust encryption and secure handling. Secondly, \emph{robustness} ensures models perform reliably across diverse conditions, minimizing diagnostic errors. \emph{Reliability} is crucial for consistent, accurate results, avoiding hallucinations and building clinician confidence in the adoption of these tools. \emph{Explainability} is needed for healthcare professionals to understand and validate the model's outputs, enhancing trust and safety alignment. \emph{Fairness} ensures equitable treatment across diverse patient groups, preventing bias and promoting ethical healthcare practices. Although these five perspectives do not cover every trustworthiness concern, they are representative enough to motivate us to survey them. We aim to provide a comprehensive trustworthiness framework that ensures medical foundation models effectively improve patient care and clinical decision-making by addressing these key issues.

\subsection{Privacy (T1)}
Privacy in medical image analysis is critical as healthcare service providers are expected to follow necessary safety measures to safeguard patients' private information, such as age and gender. This privacy concern can also exist in medical foundation models due to the presence of health information in the training dataset for medical foundation model training~\cite{brown2022does, adnan2022federated}. 
With the increase in the amount of data used for training and the complexity of foundation models, there is a growing interest in exploring privacy-preserving techniques in medical foundation models in both academia and industry.

\subsection{Robustness (T2)}
Robustness in medical image analysis refers to the ability to maintain performance when faced with various uncertainties and adverse conditions. Similarly, foundation models for medical image analysis also face challenges related to robustness concerns. This includes handling errors or imbalance in input text prompts~\cite{zhang2023segment}, being generalizable to data from various distributions~\cite{wornow2023shaky}, defending against adversarial attacks designed to mislead the model~\cite{jin2024backdoor}, and resisting attempts at data poisoning where malicious entities might attempt to influence the model's training data~\cite{wang2023decodingtrust}. Thus, the emphasis on robustness in foundation models for medical image analysis is crucial given the sensitive nature of medical diagnostics and the potential impact on patient care, requiring that these models have not only good performance but also resilient to a wide range of potential vulnerabilities.

\subsection{Reliability (T3)}
Healthcare is commonly seen as a high-stakes field, where reliability is a foundational requirement.
\textit{Firstly}, it is common that some foundation models such as LLMs can provide untruthful answers or generate misleading information, which may cause significant consequences in some scenarios such as Medical Q\&A~\cite{zhao2023chatcad}. This is particularly concerning in healthcare, where the accuracy of the information can directly impact patient outcomes~\cite{ahmad2023creating}. \textit{Secondly}, hallucination can also happen when LLM's confidence is miscalibrated. In a medical context, such overconfidence in wrong information can be dangerous. For instance, an LLM might generate a very confident but incorrect interpretation of a patient's symptoms, leading healthcare providers down the wrong diagnostic path~\cite{nori2023capabilities}.

\subsection{Explainability (T4)}
The explainability of foundation models refers to the ability to understand and interpret how these models make decisions or generate outputs~\cite{zhao2024explainability,luo2024understanding}. This property becomes crucial in medical image analysis, especially when applying foundation models in healthcare, due to the demand for trustworthy and actionable decision-making in clinical settings. To be specific, one common example of explainability in medical image analysis is highlighting the regions of an image that the model considers most indicative of a particular diagnosis. Another example involves delineating the boundaries of different anatomical structures or pathological regions, allowing practitioners to understand why certain areas were identified as significant.

\subsection{Fairness (T5)}
Fairness in foundation models refers to the equitable performance of the model across different demographic groups, ensuring that no group experiences significantly lower performance. Similarly, it is also significant to eliminate fairness issues when adapting these foundation models to medical image analysis fields~\cite{ricci2022addressing}. Due to the common issue of under-representation or imbalance in medical data, fairness in foundation models for medical image analysis is particularly critical. Therefore, it is essential to ensure that these medical foundation models perform consistently well for all populations, mitigating disparate impacts and reducing performance gaps between groups to maintain fairness.

Drawing from the above various perspectives, trustworthiness in medical image analysis tasks spans a wide range of targets, often dependent on context and application. Additionally, the key concerns about trustworthiness relate to how foundation models are adapted for specific medical imaging applications. There is an urgent need for a systematic review and discussion on the trustworthiness of foundation models in medical image analysis tasks.

\begin{table*}
\begin{threeparttable}
\centering

\resizebox{0.9\textwidth}{!}{
\begin{tabular}
{ccp{0.1\textwidth}cccp{0.28\textwidth}p{0.14\textwidth}}
\toprule[1pt]
\textbf{Applications} & \textbf{Ref.} &\textbf{FMs} & \textbf{Usage$^\dagger$} & \textbf{Trust.$^\ddagger$} & \textbf{Modalities$^\ast$} &\textbf{Datasets} & \textbf{Body Parts / Organs$^{\ast\ast}$} \\ \hline

Segmentation & \cite{xu2023fairness} & MedSAM & OS & T5 & M2 & KiTS19 & B14 \\ 
\hline
Segmentation & \cite{liu2024fedfms} & FedSAM & FT & T1 & M3, M7 & PROMISE12, FeTS2022, PanNuke, MoNuSAC2020, G1020, Origa-light & B7, B18, B34\\ 
\hline
Segmentation & \cite{asokan2024federatedlearningfriendlyapproachparameterefficient} & SAM & FT & T1 & M2, M3 & KiTS19, Fed-IXI, Prostate MRI & B14, B18, B34\\ 
\hline
Segmentation & 
\cite{huang2024segment} & SAM & OS & T2 & M1-M3, M6, M8-M10 & COSMOS 1050K (combination of 53 sets) & B1-B7, B12, B16, B17 \\ 
\hline
Segmentation & \cite{zhang2023segment} & SAM & PT & T2 & M2 & StructSeg, FLARE 22 & B2-B4, B7, B8 \\ 
\hline
Segmentation & \cite{wang2023sam} & SAM & PT & T2 & M7 & EndoVis17, EndoVis18 & B15 \\ 
\hline
Segmentation & \cite{wong2023scribbleprompt} & SAM & PT & T2 & M2, M3, M6, M11 & 65 external sets & B10, B18-B21 \\ 
\hline
Segmentation & \cite{roy2023sammd} & SAM & PT & T2 & M2 & AMOS22 & B14, B22-B32 \\ 
\hline
Segmentation & \cite{stein2023influence} & SAM & PT & T2 & M3 & ACDC, M\&Ms & B33 \\
\hline
Segmentation & \cite{jiang2024uncertaintyaware} & UA-SAM & FT & T3 & M2, M7 & LIDC-IDRI, REFUGE2 & B7, B20\\ 
\hline
Segmentation & \cite{lei2023medlsam} & MedLSAM & PT & T2 & M2 & 16 separate datasets & B1-B3\\ 
\hline
Segmentation & \cite{zhang2023segment} & UR-SAM & TS & T2 & M2 & StructSeg, FLARE 22 & B2-B4, B7, B14, B24, B26, B33 \\ 
\hline
Report Generation & \cite{zhong2023chatradiovaluer} & LLaMA-2, GPT-4 & FT, PT & T2 & M2, M3 & internal data & B1-B4 \\ 
\hline
Report Generation & \cite{chen2023s4mgeneratingradiologyreports} & MedCLIP & FT & T2 & M1 & IU-Xray, private data & B1, B4, B13, B35-37 \\ 
\hline
Report Generation & \cite{zhang2023samguidedenhancedfinegrainedencoding} & CLIP & FT & T2 & M1-M4, M6 & COCO, ROCO, MedICaT & - \\ 
\hline
Report Generation & \cite{yang2023customizing} & EVA-ViT-g, ChatGLM-6B & FT, PT & T3 & M1-M4, M6 & ImageCLEFmedical 2023 & B1-B13, B16, B17 \\ 
\hline
Report Generation & \cite{hyland2023maira1} & Vicuna & FT & T3 & M1 & MIMIC-CXR & B1 \\ 
\hline
Report Generation & \cite{tanno2023consensus} & Flamingo-CXR & FT & T3 & M1 & MIMIC-CXR, IND1 & B1 \\ 
\hline
Report Generation & \cite{lee2023llmcxr} & Dolly & FT & T3 & M1 & MIMIC-CXR & B1 \\ 
\hline
Report Generation & \cite{ramesh2022improvingradiologyreportgeneration} & GPT-3, BioBERT & OS, PT & T3 & M1 & MIMIC-CXR & B1 \\ 
\hline
Report Generation & \cite{li2023enhancedknowledgeinjectionradiology} & ClinicalBERT & OS & T3 & M1 & IU-Xray, MIMIC-CXR & B1 \\ 
\hline
Report Generation & \cite{danu2023generation} & RadBloomz & FT, PT & T4 & M1 & MIMIC-CXR & B1 \\ 
\hline
Report Generation & \cite{chen2024finegrainedimagetextalignmentmedical} & Dolly & FT & T4 & M1 & MIMIC-CXR, OpenI & B1 \\ 
\hline
Report Generation & \cite{Tanida_2023} & GPT-2 & FT & T4 & M1 & Chest ImaGenome & B1 \\ 
\hline
Medical Q\&A & \cite{chen2024chexagent} & Mistral-7B & FT & T5 & M1 & Wikipedia, CheXinstruct, MIMIC-CXR, PadChest, BIMCV-COVID-19, PMC Article, MIMIC-IV & B1 \\ 
\hline
Medical Q\&A & \cite{zhang2023pmcvqa} & LLaMA, CLIP & FT & T5 & M1-M4, M6 & PMC-VQA, VQA-RAD, SLAKE & B1, B2, B5\\ 
\hline
Medical Q\&A & \cite{vansonsbeek2023openended} & CLIP, GPT-2 & FT, PT & T2 & M1-M3, M5 & SLAKE, PathVQA, OVQA & B1-B5, B11, B12 \\ 

\bottomrule
\end{tabular}}
\vspace{-20pt}
\end{threeparttable}

\end{table*}

\begin{table*}
\begin{threeparttable}
\centering

\resizebox{0.9\textwidth}{!}{
\begin{tabular}
{ccp{0.16\textwidth}cccp{0.28\textwidth}p{0.14\textwidth}}
\toprule[1pt]
\textbf{Applications} & \textbf{Ref.} &\textbf{FMs} & \textbf{Usage$^\dagger$} & \textbf{Trust.$^\ddagger$} & \textbf{Modalities$^\ast$} &\textbf{Datasets} & \textbf{Body Parts / Organs$^{\ast\ast}$} \\ \hline

Medical Q\&A & \cite{kim2023rollama} & RO-LMM & FT, PT & T2 & M2, M3, M6 & internal data & B1 \\ \hline

Medical Q\&A & \cite{eslami2021does} & PubMedCLIP & FT & T2 & M1-M3 & VQA-RAD, SLAKE & B1-B5 \\ 
\hline
Medical Q\&A & \cite{wang2023chatcad} & ChatGPT & OS, PT & T3 & M1 & MIMIC-CXR, CheXpert & B1 \\ 
\hline
Medical Q\&A & \cite{zhao2023chatcad} & ChatGPT, CLIP & OS, PT & T3 & M1, M3 & MIMIC-CXR, private data & B1, B10, B13 \\ 
\hline
Medical Q\&A & \cite{thawkar2023xraygpt} & MedCLIP, Vicuna & PT & T3 & M1 &  MIMIC-CXR & B1 \\ 
\hline
Medical Q\&A & \cite{shaaban2024medpromptx} & Med-Flamingo & OS, PT & T3 & M1 & MedPromptX-VQA & B1 \\ 
\hline
Medical Q\&A & \cite{wang2022medclip} & MedCLIP & TS, FT & T2 & M1 & MIMIC-CXR, CheXpert, COVID, RSNA Pneumonia & B1 \\ 
\hline
Disease Diagnosis & \cite{liu2023chatgpt} & ChatGPT, CLIP & OS & T4, T5 & M1, M3, M7 & Pneumonia, Montgomery, Shenzhen, IDRID, BrainTumor & B1, B2, B7, B18 \\
\hline
Disease Diagnosis & \cite{wu2023can} & GPT-4 & OS & T3 & --
& -- & B1-B13\\ \hline

Disease Diagnosis & \cite{yang2024demographic} & CheXzero & OS & T5 & M1 & MIMIC, NIH, CheXpert, PadChest, VinDr & B1 \\ \hline
Disease Diagnosis & \cite{luo2024fairclip} & CLIP, BLIP2 & OS & T5 & M12 & Harvard-FairVLMed & B7 \\ \hline
Disease Diagnosis & \cite{jin2024backdoor} & MedCLIP & FT & T2 & M1 & MIMIC, COVIDX, RSNA & B1 \\ \hline
Disease Diagnosis & \cite{berrada2023unlocking} & NFNet & FT & T1, T5 & M1 & MIMIC-CXR, CheXpert & B1 \\ \hline
Disease Diagnosis & \cite{bie2024xcoop} & CLIP & PT & T4 & M1, M2, M9 & SkinCon, Pneumonia, IU X-Ray & B1, B6, B7 \\ \hline
Disease Diagnosis & \cite{agarwal2023representing} & CLIP & OS & T4 & M5, M9 & CBIS-DDSM, SIIM-ISIC & B1, B6 \\ \hline
Disease Diagnosis & \cite{doerrich2024integrating} & DINOv2 & OS & T1, T4 & M1, M9 & Pneumonia, Melanoma & B1, B6 \\ \hline
Disease Diagnosis & \cite{yan2023robust} & GPT-4 & OS & T4 & M1 & NIH-CXR, Covid-QU, Pneumonia, Open-i & B1 \\ \hline
Disease Diagnosis & \cite{wei2024visionclip} & VisionCLIP & TS & T1 & M7 & MESSIDOR, FIVES, REFUGE & B7 \\ 

\bottomrule
\end{tabular}}
\caption{\textbf{Papers of foundation models for medical image analysis.}\\
}

\begin{tablenotes}
\scriptsize{
\item $\dagger$ Training From Scratch (TS), Fine-Tuning (FT), Prompt-Tuning (PT), Off-the-Shelf (OS)
\item $\ddagger$ Privacy (T1), Robustness (T2), Reliability (T3), Explainability (T4), Fairness (T5)
\item $^\ast$ X-ray (M1), CT (M2), MRI (M3), PET/CT (M4), Histopathology (M5), Ultrasound (M6), \\Photography (M7), Endoscopy (M8), Dermoscopy (M9), Microscopy (M10), OCT (M11), \\Funduscopy (M12)
\item $^{\ast\ast}$ Chest (B1), Head (B2), Neck (B3), Abdomen (B4), Pelvic (B5), Skin (B6), Eye (B7), ENT (B8), \\Foot (B9), Tooth (B10), Hand (B11), Leg (B12), Knee (B13), Kidney (B14), Porcine Kidney (B15), \\Spine (B16), Polyp (B17), Brain (B18), Bones (B19), Thorax (B20), Cells (B21), Spleen (B22), \\Gall Bladder (B23), Esophagus (B24), Liver (B25), Stomach (B26), Aorta (B27), Postcava (B28), \\Pancreas (B29), Adrenal Gland (B30), Intestine (B31), Bladder (B32), Heart (B33), Prostate (B34), \\Hip (B35), Wrist (B36), Shoulder (B37)
}
\end{tablenotes}

\label{tab:papers}
\vspace{-20pt}
\end{threeparttable}

\end{table*}

\section{Unpacking Trustworthiness in Current Foundation Models Techniques for Medical Imaging Tasks}
\label{applications}
Table~\ref{tab:papers} lists the publications reviewed in this study on the development of medical imaging foundation models, specifically highlighting those that include explicit discussions or evidence regarding their trustworthiness.  If several foundation models are used in a paper, the two most representative ones are listed. In addition, the usage of the foundation model is categorized into the four medical imaging applications we focus on, as well as the trustworthiness issues. All the modalities, datasets, and body parts or organs covered in each literature are also identified. Through this structured approach, readers can easily navigate the landscape of current research on medical foundation models and identify key trends and gaps in trustworthiness.

\subsection{Segmentation}
Image segmentation is a crucial task in medical image analysis. A substantial stride in the development of a foundation model for medical image segmentation came with the introduction of the Segment Anything Model (SAM)~\cite{kirillov2023segment}. SAM is a vision transformer (ViT) model developed by Meta AI which demonstrates great zero-shot performance on diverse sets of natural images. Inspired by the generalizability gained through prompt engineering of LLMs, SAM was designed to accept prompts as points, boxes, or text. For now, only point and box prompts are publicly available. These prompts are encoded using a CLIP encoder. Based on SAM, several models, modifications and frameworks have been developed for medical image analysis, including MedSAM \cite{ma2024segment}, MedLSAM \cite{lei2023medlsam}, UR-SAM \cite{zhang2023segment}, UA-SAM \cite{jiang2024uncertaintyaware}, FedSAM \cite{liu2024fedfms}, ScribblePrompt \cite{wong2023scribbleprompt}, and GazeSAM \cite{wang2023gazesam}.

\subsubsection{Privacy in Segmentation}

\noindent\textbf{Medical image de-identification and federated learning have been adopted in SAM-based models for privacy preservation} Privacy concerns in foundation models for medical image segmentation represent a critical challenge, given the sensitive nature of patient data processed by these models. Most foundation models for medical image segmentation are based on SAM architecture~\cite{kirillov2023segment}, which asserts that the 11 M natural image dataset they use is ``privacy respecting'' - which they specify as meaning that faces and license plates are blurred out. Similarly, in MedSAM \cite{ma2024segment} the authors leveraged the de-identified public medical images for model training. Additionally, federated learning, a method where model training is done across multiple protected data sources without sharing data~\cite{cheng2020federated}, have been adopted to train or finetune SAM-based models. It avoids the patient confidentiality concerns of storing and sharing privacy-sensitive medical data  Liu~\etal~\cite{liu2024fedfms} presents a framework, named FedSAM, for federated SAM fine-tuing across multiple clients, each of which has access to a fraction of the total dataset. Using prostate cancer MRI, brain tumour MRI, nuclei slide images, and fundus photograph sets for training, FedSAM shows comparable performance to SAM, except with the nuclei slide images, where the dataset size is the smallest \cite{liu2024fedfms}. Similarly, Asokan~\etal introduces a method for identifying performance-important layers of a fine-tuned SAM, thus enabling parameter-efficient federated learning, resulting in 6\% higher Dice score for a SAM fine-tuned by federated learning on the KiTS19 kidney CT scan dataset \cite{asokan2024federatedlearningfriendlyapproachparameterefficient}.

\subsubsection{Robustness in Segmentation}

Robustness in foundation models for medical image segmentation is crucial for consistent and accurate results across various conditions. It ensures reliable outcomes, regardless of changes in data input or computational environment. This issue is particularly vulnerable in SAM-based segmentation models, where users can input customized prompts, potentially leading to inconsistent results due to the variability and unpredictability of these user-defined inputs. \\

\noindent\textbf{Robustness issues across different image modalities.} 
SAM and its variants have claimed to achieve superior performance in image segmentaion, including medical images. Is it a cure-all tool for widespread medical image modalities? Unfortunatley, existing studies show that they still suffer from  a lower tolerance for uncertainty. A study into the performance of the base SAM on a collection of 556k medical image-mask pairs covering 16 image modalities by Huang~\etal noted great performance on some modalities, but partial or complete failure on other image modalities. They state that the performance difference between modalities indicates that SAM ``cannot stably and accurately implement zero-shot segmentation on multimodal and multi-object medical datasets'' \cite{huang2024segment}. \textit{To address this, several methods related to uncertainty estimation are proposed to mitigate the robustness issue.} Zhang~\etal introduces uncertainty estimation for both SAM and MedSAM by using the change in segmentation boundaries as a function of prompt augmentation to generate uncertainty maps \cite{zhang2023segment}. They propose that incorporating uncertainty estimations into SAMs builds trust through better error identification. Jiang~\etal notes that features in medical images may have ambiguous boundaries, in contrast to the clear boundaries of most natural images. They propose UA-SAM~\cite{jiang2024uncertaintyaware}, which integrates a probabilistic model into SAM training and fine-tuning.
In addition to fine-tuning the SAM for domains with inherent uncertainty, the adapter in UA-SAM makes the SAM non-deterministic by outputting multiple ``plausible'' masks for a single input. \\

\noindent\textbf{Robustness issues on mid-surgery images with blur, reflections, or other types of noise.} Images taken during surgery itself represent an important use case for segmentation. These kinds of images differ in image quality from the natural images used to train segmentation foundation models like SAM.  Wang~\etal qualitatively assesses the performance of a base SAM on mid-surgery images, and finds that image features like blur or reflections result in inaccurate segmentation \cite{wang2023sam}. This is concerning as those features would be found throughout the surgery process. Additionally, Wang~\etal quantitatively tests the performance of SAM on mid-surgery images from the EndoVis17 image set \cite{allan20192017}, while imposing 18 different types of noise (brightness, defocus, saturation, JPEG compression etc.) with different severity levels~\cite{wang2023sam}. Some image corruptions, like saturation, only result in a slight decrease in performance, while others, like JPEG compression, cause near total failure~\cite{wang2023sam}. \\

\noindent\textbf{Robustness issues on using  different prompt inputs.} As mentioned, many SAM-based medical image segmentation models accept two types of prompts: point prompts and box prompts. Prompting gives information to the model on what parts of the image are useful and where boundaries are. This allows a user to contribute to delineating a boundary that the model may not initially detect. This has relevance in the context of medical image segmentation in that the low contrast of medical images (as opposed to the SAM training set of natural images) often requires prompting to get to acceptable performance levels for clinical tasks. \\
\noindent\textbf{Automatic prompting methods are proposed, but the performance is not as good as manual prompting}  Lei~\etal introduces MedLSAM, a foundation model trained on 14,012 CT scan images. MedLSAM can automatically generate box prompts. Evaluating on two sets of CT scan images covering 48 anatomical regions or organs, Lei~\etal finds that auto-prompting performance is generally worse than manual prompting, but that for some anatomical features the performance is comparable~\cite{lei2023medlsam}. Addressing MedLSAM, Zhang~\etal proposes automatic prompt generation for SAMs on medical images, a framework they call UR-SAM~\cite{zhang2023segment}. UR-SAM perturbs prompts to generate an uncertainty map of potential box prompts. UR-SAM achieves better performance using uncertainty maps than base auto-prompting, though performance is still not as good as manual prompting \cite{zhang2023segment}. \\

\noindent\textbf{The robustness testing on using positive point prompts and box prompts} The point prompts that SAMs can accept can be positive or negative (containing or not containing the target, respectively). An analysis by Stein~\etal gauges the performance benefits of different types of prompts for a base SAM on a set of cardiac magnetic resonance imaging (cMRI) images for segmentation.  Roy~\etal perform box-prompt segmentation on a collection of CT scan images covering 14 organs while incorporating jitter. They perform the robustness testing by simulating the errors made by a user, for example, applying different magnitutes jitter to the box prompts. They find that jitters less than and including 0.1 only marginally decrease performance, whereas jitters of 0.25 and 0.5 cause extremely poor performance. This indicates that some amount of uncertainty in box-prompting can be tolerated by SAM for medical images \cite{roy2023sammd}. Additionally, Roy~\etal compares using different number of points (\textit{i.e.}, 1 point, 3 point, and 10 points prompts) on a multi-organ CT scan image set. Unsurprisingly, the more points that are used, the better the performance. Finally, they compare box prompts to point prompts, and find that box prompts perform considerably better than even the 10 point prompts \cite{roy2023sammd}. They find benefits to using positive point prompts, and further benefit in using positive point prompts in conjunction with a box prompt \cite{stein2023influence}. Interestingly, using positive and negative point prompts in conjunction with a box prompt results in little-to-no improvement over the unprompted SAM.

Literature in SAM-based medical image segmentation has also explored other types of prompts, including scribble-based~\cite{wong2023scribbleprompt}, gaze-bsed~\cite{wang2023gazesam}, and text-based~\cite{zhao2024biomedparse}, though robustness discussion is not covered.

\subsubsection{Fairness in Segmentation}
\noindent\textbf{Imbalanced training set and variability of clinician needs in segmentation tasks may cause fairness issues.} As with base SAM itself, limitations are identified in the ability of MedSAM to perform reliably with underrepresented image modalities. In the context of MedSAM and medical image analysis, the concern comes from the fact that the training set largely consists of CT, MRI, or Endoscopy images. A similar issue is found in the variability of clinician needs in segmentation tasks; not all image modalities should be divided the same way. This presents a concern in the \emph{fairness} of these segmentation models as they are used in rare clinical scenarios \cite{ma2024segment}.\\

\noindent\textbf{Performance disparities related to patient BMI and gender are observed in MedSAM.} MedSAM~\cite{ma2024segment}, a customization of the SAM fine-tuned on over 1M medical image-mask pairs covering 15 image modalities, has become a popular medical image analysis foundation model for segmentation.
Xu~\etal\cite{xu2023fairness} examines the performance of MedSAM on a kidney CT scan set and looks for disparities in performance depending on patient characteristics. They use Dice similarity scores as their performance metric. No statistically significant relationship between patient age and performance is found, but a statistically significant negative correlation between patient BMI and performance has been noted. Xu~\etal\cite{xu2023fairness} also find that images from female patients yield better performance than from male patients. 

\subsection{Report Generation}

The adoption of foundation models in the generation of medical reports represents a pivotal advancement, facilitating the creation of patient reports and radiology interpretations. However, this technological leap has also brought to light trustworthiness concerns, including data privacy, the reliability of generated reports, and the necessity for transparent, explainable models. This section explores potential trustworthiness issues and solutions presented in the literature.

\subsubsection{Privacy in Report Generation}
\noindent\textbf{Deidentication on both medical image data and text reports.}
Medical applications inherently come with significant concerns regarding patient data privacy. There are standard pipelines to remove sensitive informations from the medical image DICOM files, but how to provent  the reports leaking sentitive patient privacy remains lacking standard guidelines.
To satisfy privacy requirements, Thawkar~\etal\cite{thawkar2023xraygpt} use the MIMIC-CXR dataset where the DICOM files of the X-ray data samples are de-identified by the data provider. They further by removing patient information when training their model. To preserve the utility of the model trained on de-identified data, the authors used GPT-3.5-turbo to remove the de-defined symbols ``\_\_'', while preserving the original meaning of the reports.

\subsubsection{Robustness in Report Generation}
Medical report generation is a high-risk field with a low tolerance for errors, making it especially important to ensure the robustness of medical foundation models.
However, due to the complexity and variance across sources like different inspected body regions and institutions, ensuring the robustness in medial foundation models still faces numerous challenges.\\

\noindent\textbf{Robustness issues on applying to diverse data types and sources.}
To improve the robustness of these models across different body regions, Zhong~\etal \cite{zhong2023chatradiovaluer} propose ChatRadio-Valuer, a model based on LLMs that learns generalizable representations from radiology reports of one institution and adapts to report generation tasks across different body regions (chest, abdomen, muscle-skeleton, head, maxillofacial, and neck). Another work on body-part robustness in report generation proposes a single-for-multiple (S4M) framework to make the model robust to six different body parts (chest, abdomen, knee, hip, wrist, and shoulder) while maintaining its performance or even outperforming other baseline methods~\cite{chen2023s4mgeneratingradiologyreports}. This is achieved by incorporating general radiology knowledge with the radiology-informed knowledge aggregation branch and enhancing the cross-modal alignment by the implicit prior guidance branch.
Another robustness issue comes from the heterogeneity of different clinical institutions.
To solve this problem, \cite{zhong2023chatradiovaluer} starts with training on radiology reports from one institution to acquire knowledge of particular patterns and representations, and then it undergoes supervised fine-tuning using data from other numerous institutions. This approach ensures that the model can effectively handle a wide range of clinical scenarios and diagnostic tasks, improving its generalization ability and robustness to different institutions in real-world applications.\\

\noindent\textbf{Robustness to poor data quality.}
Medical images with blurry boundaries or noise can also lead to robustness concerns. To address the unstable generalization issues, an innovative SAM-guided dual-encoder architecture in MSMedCap is utilized to enable the capture of information with different granularities and overcome the unstable performance due to blurry boundaries, noise, and poor contrast in medical images~\cite{zhang2023samguidedenhancedfinegrainedencoding}. 
Inspired by NEFTune~\cite{jain2023neftune}, RO-LLaMA also enhances the model’s robustness and generalization abilities when faced with noisy inputs by adding random noise to the embedding vectors during training~\cite{kim2023rollama}.

\subsubsection{Reliability in Report Generation}
Inheriting the issue of general foundation models, for example LLMs and VLMs, report generation systems using foundation models also suffer from reliability issues.\\

\noindent\textbf{Removing prior references from medical reports to improve reliability.} Medical foundation models can make hallucinated references to non-existent prior reports when generating medical reports. 
One reason for this is that these models are trained on datasets of real-world patient reports that inherently refer to prior reports and lead to the incompatibility when solely inputting a medical image~\cite{ramesh2022improvingradiologyreportgeneration, hyland2023maira1}. To address this, \cite{ramesh2022improvingradiologyreportgeneration} propose CXR-ReDonE to remove prior references from medical reports by using GPT-3 to rewrite the reports and using token classification to remove words referring to priors.\\

\noindent\textbf{Improving training data, LLMs, or injecting additional knowledge to improve reliability.} In addition, Lee~\etal\cite{lee2023llmcxr} find that their proposed method for chest X-ray report generation may hallucinate reports with nonexistent findings. To mitigate this issue, they suggest some potential solutions, such as using higher quality/quantity training data, larger LLMs, or to strengthen the alignment between image and text modalities. Another work~\cite{li2023enhancedknowledgeinjectionradiology} proposes to enhance the reliability of radiology report generation by injecting additional knowledge with the current image in the generation process. Specifically, they fuse the weighted medical concept knowledge and reports for similar images with the current image features. Another study observe that when customizing general-purpose foundation models for medical report generation, they can hallucinate unintended text~\cite{yang2023customizing}. They attribute this to the use of cross-entropy loss for language modeling, which leads to the problem of exposure bias. One potential solution to address this issue is to leverage reinforcement learning with human (AI) feedback.

\subsubsection{Explainability in Report Generation}

\noindent\textbf{Using bounding boxes and LLMs to improve explainability.} Explainability is also a crucial concern in medical report generation. To improve the interpretability of report generation systems, Danu~\etal \cite{danu2023generation} propose a two-step approach for generating the findings section of a radiology report from an automated interpretation of chest X-ray images: (1) detecting the abnormalities within the image using bounding-boxes with probabilities, and (2) harnessing the power of LLMs to translate the list of abnormalities into a Findings report. This two-step approach adds interpretability to the framework and aligns it with radiologists' systematic reasoning during the review of CXRs.\\

\noindent\textbf{Using adaptive patch-word matching to improve explainability.} Using fine-grained vision-language models to provide alignment between image patches and texts which can match parts of the generated report with specific regions in the medical image can also improve explainability. However, using image patches with fixed sizes may result in incomplete representation of lesions which can occur in varying sizes. To address this issue, Chen~\etal\cite{chen2024finegrainedimagetextalignmentmedical} propose an Adaptive patch-word Matching (AdaMatch) model which matches texts from the report with adaptive patches in the medical image to provide explainability. It utilizes an Adaptive Patch extraction module to dynamically capture abnormal regions of varying sizes and positions. To further improve its explainability, they propose a cyclic CXR-report generation pipeline (AdaMatch-Cyclic), which can perform both CXR-to-report and report-to-CXR generation based on patch-word alignment.\\

\noindent\textbf{Linking report texts to regions in the medical image to improve explainability.} Tanida~\etal\cite{Tanida_2023} propose a region-guided radiology report generation model that describes individual regions to form the final report. Specifically, an object detector is used to identify and extract visual features of distinct anatomical regions in the chest X-ray image, and then the abnormal regions are selected and passed to a pre-trained LLM like GPT-2 which generates sentences for each selected region. Finally, the generated sentences are post-processed to remove duplicates and concatenated to form the final report. This process ensures that each sentence in the final report can be linked to an abnormal region in the medical image, which greatly improves explainability.

\subsection{Medical Q\&A}
With the power of language and vision foundation models, Medical Q\&A systems leverage their ability for image and text understanding to serve for a wide range of tasks~\cite{wang2023chatcad, zhao2023chatcad,thawkar2023xraygpt,wang2023gptdoctor}. These Medical Q\&A systems usually accept a medical image and a text prompt as inputs and output text in response to the question. 
However, similar to medical report generation tasks, the appearance of LLMs in the medical domain has also accentuated concerns regarding their trustworthiness. Fortunately, the research community has begun identifying and addressing these concerns as well as proposing methods to mitigate the trustworthiness issues.

\subsubsection{Robustness in Medical Q\&A}
Robustness in medical foundation models for Q\&A tasks requires being  
able to handle errors in the input text prompts as well as be resilient against adversarial and backdoor attacks in medical images.\\

\noindent\textbf{Foundation models can be sensitive to prompt structures.} Though some medical foundation models achieve success in the model performance, the susceptibility of these models to specific text inputs still poses a significant challenge.
Sonsbeek~\etal~\cite{vansonsbeek2023openended} proposes a prefix tuning method for open-ended medical VQA of LLMs for performance improvement, but the experimental results show that the model is heavily sensitive to the prompt structure, as swapping the order of the question embeddings and the visual prefix yields a decrease in performance.\\

\noindent\textbf{Consistency regularization can be used to improve model's robustness against noisy inputs.} To improve the robustness of LLM-based foundation models in medical Q\&A, RO-LLaMA~\cite{kim2023rollama} incorporates consistency regularization, inspired by NEFTune~\cite{jain2023neftune}, which adds random noise to the embedding vectors during training. This approach enhances the model's robustness and generalization abilities when faced with noisy inputs.\\

\noindent\textbf{CLIP-based models are susceptible to backdoor attacks.} Additionally, robustness issues are also present in CLIP-based medical Q\&A models such as MedCLIP~\cite{wang2022medclip} and PubMedCLIP~\cite{eslami2021does}, especially if the backbone CLIP-based models themselves have robustness problems. This concern can be inferred by observations in~\cite{jin2024backdoor}, which show that using a modest set of wrongly labeled data and introducing a ``Bad-Distance'' between the embeddings of clean and poisoned data can lead to successful backdoor attacks.

\subsubsection{Reliability in Medical Q\&A}
The reliability of foundation models is a significant concern in Q\&A domains, particularly because these models can produce untruthful answers or hallucinations. This challenge is further complicated by the fact that these models often inherit the same issues when adapted to other fields. Unlike general applications, where some degree of unreliability may be tolerated, reliability in the medical domain is crucial. Unreliable responses in medical Q\&A can disastrously mislead decision-making, potentially resulting in harmful consequences~\cite{chen2024detecting}.\\

\noindent\textbf{Benchmarks for Hallucinations.} 
One significant threat for reliability of medical foundation models is hallucination, which is hard to evaluate and detect. Although existing works like CHAIR~\cite{rohrbach2018object} are designed for general foundation hallucination evaluation, they cannot be directly adapted to the medical domain. To address this issue, the first benchmark, Med-HallMark, and a new evaluation metric, MediHall Score, are proposed for hallucination detection specific to medical foundation models, providing baselines for various models in medical image analysis~\cite{chen2024detecting}. Meanwhile, a hallucination detector for medical foundation models is proposed for five types of hallucinations: catastrophic, critical, attribute, prompt-induced and minor hallucinations. Similarly, CARES is proposed to comprehensively evaluate the hallucination of Medical Large Vision Language Models (Med-LVLMs)~\cite{xia2024cares}. To ensure reliability, Shaaban~\etal~\cite{shaaban2024medpromptx} propose MedPromptX and leverage few-shot prompting (FP) to eliminate hallucination in the foundation model for chest X-ray Diagnosis by guiding the output.\\

\noindent\textbf{Potential issues and solutions of truthfulness.} Another concern in reliability is truthfulness. Retrieving knowledge from professional sources can improve the truthfulness of LLMs' answers. A recent study highlights that Medical Visual Question Answering (Med-VQA) systems may be unreliable for medical diagnosis questions and could produce misleading information when existing state-of-the-art foundation models, such as GPT-4, are directly adapted to medical image analysis domains~\cite{yan2024worse}. An example of this is ChatCAD, which leverages LLMs' extensive medical knowledge to provide interactive explanations and advice~\cite{wang2023chatcad}. However, the truthfulness of ChatCAD is threatened by the restricted scope of applicable imaging domains and the lack of requisite depth in medical expertise during interactive patient consultations. To address these issues, Zhao~\etal~\cite{zhao2023chatcad} propose ChatCAD+, which incorporates a domain identification module to work with a variety of CAD models. Additionally, outdated medical information can also contribute to incorrect answers. Instead of directly answering medical questions, Retrieval-Augmented Generated (RAG) is utilized in ChatCAD+ for obtaining relevant, up-to-date information, which further enhances the truthfulness of its answers. 

\subsubsection{Explainability in Medical Q\&A}
\noindent\textbf{LLMs can be used to provide explainations.} In high-risk medical environments, the use of black-box foundation models can lead to severe consequences, which may directly impact patient health and safety. Therefore, it is essential to ensure that the decision-making processes of medical foundation models are transparent and explainable. By providing interactive explanations for the advice of the medical image, ChatCAD enhances its explainability for clinical decision-making~\cite{wang2023chatcad}. 
With the popular use of GPT-4 in medical fields, providing appropriate prompts that ask the model to explain its answers during medical Q\&A can also enhance the explainability~\cite{nori2023capabilities}.

\subsubsection{Fairness in Medical Q\&A}
Under-representation of certain groups in the training data or training data imbalance can lead to significantly lower accuracy or effectiveness of the model for certain populations, leading to disparate performance and fairness issues \cite{liu2023trustworthy}. 
In medical Q\&A systems, these fairness issues can be caused by imbalanced training data in both the medical image and language domains. \\

\begin{figure*}[h]
    \centering
    \includegraphics[width=0.95\linewidth]{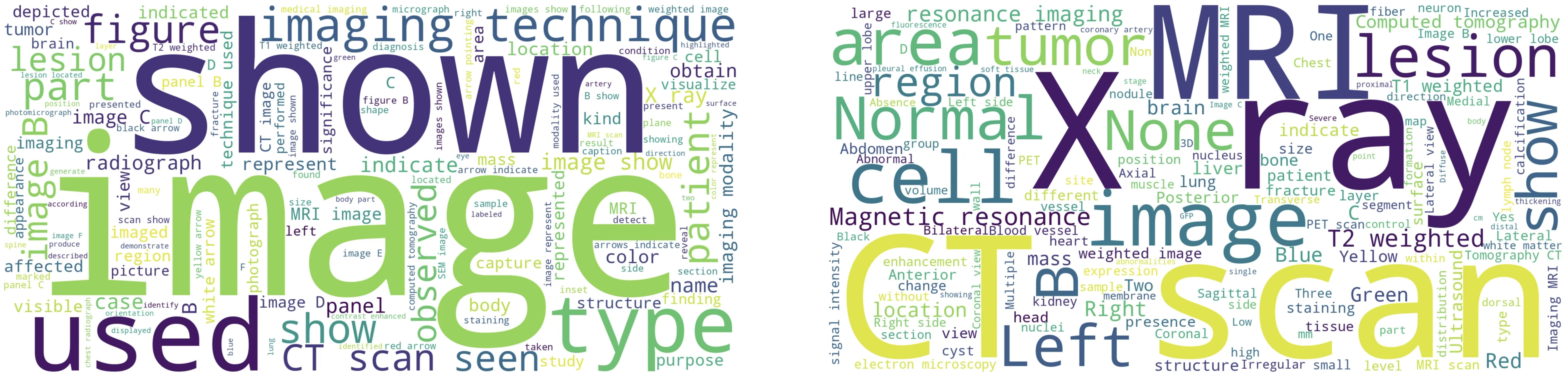}
    \caption{Word cloud visualization of questions (left) and answers (right) from the training set of the PMC-VQA dataset.}
    \label{fig: wordcloud}
\end{figure*}

\noindent\textbf{Fairness issues can arise from data collection, annotation, or the distribution of the dataset.} Zhang ~\etal \cite{zhang2023pmcvqa} propose PMC-VQA, a large-scale medical VQA dataset with 227k VQA pairs of 149k images covering various modalities and diseases. Despite their efforts to construct a comprehensive MedVQA dataset, the authors note the potential presence of biases, which may arise from data collection, annotation (inconsistencies or subjective interpretations from human annotators), or the underlying distribution of the medical images and questions. Figure \ref{fig: wordcloud} shows the word clouds for the questions and answers in the training set of the PMC-VQA dataset.
The imbalance in word frequencies may be an indication of potential biases in the dataset. For instance, for the word cloud of answers, ``X-ray", ``MRI", and ``CT scan" are the most prevalent words, suggesting that models trained on this dataset may perform better on well-represented modalities like X-ray, CT, and MRI, and worse on other imaging modalities.\\

\noindent\textbf{Fairness issues could arise from imbalanced training data such as SLAKE and CheXpert.} Another popular dataset used for medical Q\&A tasks is the SLAKE \cite{liu2021slake} dataset. 
The distribution of images for different body parts in SLAKE dataset consists mostly of images of chest, abdomen, and head, with pelvic and neck images only taking up a significantly smaller portion. 
Thus, the under-representation of pelvic and neck images may
lead to worse performance when tested on images from these body regions. Another example is the CheXpert dataset, which is a large public dataset of chest radiographs of 65,240 patients. The age distribution of male and female patients reveals a gender imbalance in the dataset, with a predominance of male patients for most of the age groups, potentially leading to fairness issues in which the model performs better at diagnosing radiographs from male patients.\\

\noindent\textbf{Performance disparities are observed in a foundation model designed for diverse group of datasets.} Chen \etal \cite{chen2024chexagent} propose CheXagent, a foundation model designed for chest X-ray interpretation, consisting of three main components: a clinical LLM, a vision encoder, and a vision-language bridger.
To evaluate potential fairness concerns present in their proposed model, the authors tested it on a subset of the CheXpert dataset, using chest X-rays from individuals self-reporting as Asian, White, or Black. Resampling with replacement was used to ensure balanced disease prevalence and subgroup representation. The model's performance was then evaluated for cardiomegaly detection using F1-scores with the prompt “Does this chest X-ray contain cardiomegaly?" Some disparities were observed with the model's performance on patients of different races and ages. Specifically, F1-scores are highest for the Black subgroup and lowest for the Asian subgroup, reflecting potential inherent differences in the presentation of cardiomegaly across races, and could also be influenced by the limited samples of 14 Black and 30 unique Asian subjects included in the test set. Regarding age, the model performs better for the 65+ age group compared to the 0-65 group, potentially due to a higher prevalence of cardiomegaly in older patients, and age-related physiological differences. To address this fairness issue, one possible way is through curating larger and more diverse datasets, ensuring that models trained using them are more representative and equitable across different patient demographics.\\

\noindent\textbf{Stereotype bias is observed in multi-modal foundation models.} Stereotype bias also leads to fairness concerns, which are typically misleading views and expectations toward particular social groups. The most common stereotype biases include bias towards gender, race, religion, sexual orientation, disability, socio-economic status, and age \cite{liu2023trustworthy}. Most LLMs are trained on data from the internet, which is riddled with various sources of inherent biases, and LLMs may perpetuate or amplify these existing biases. For example, Bubeck \etal revealed GPT-4’s choice of pronouns amplifies the skewness of the world representation for that occupation~\cite{bubeck2023sparks}. 
Thus, the integration of LLMs, either directly or after fine-tuning, into medical Q\&A systems, requires careful consideration of these inherent bias and fairness issues of the LLMs.

\subsection{Disease Diagnosis}
Disease diagnosis, a fundamental aspect of healthcare, primarily relies on the analysis of results from diagnostic tests such as medical imaging. With the recent advancements in foundation models, these sophisticated AI systems have become increasingly beneficial in enhancing the performance and trustworthiness of disease diagnoses. By integrating foundation models, healthcare professionals can leverage deeper insights and more precise interpretations of complex diagnostic data, leading to more reliable and effective diagnoses.

\subsubsection{Privacy in Disease Diagnosis}
\noindent\textbf{Trade-off problem when using differential privacy for model fine-tuning.} Due to the fact that foundation models for medical image analysis are typically trained on large-scale datasets which may contain sensitive information from patients, privacy considerations in foundation models for medical image analysis are increasingly paramount. As a gold standard framework for preserving privacy, differential privacy (DP) can provide privacy guarantees for machine learning training without information leakage. However, degradations in model performance pose an obstacle to its application.
To tackle this challenge, a privacy-preserving approach is proposed, involving the fine-tuning of pre-trained foundation models with differential privacy (DP) \cite{berrada2023unlocking}. Numerous experimental results indicate that this approach can achieve comparable accuracy to non-private classifiers for medical image analysis, even in the presence of substantial distribution shifts between pre-training data and downstream tasks.\\

\noindent\textbf{Data distillation and data synthesis for privacy preservation.} In addition, privacy threats can also occur in the data sharing process before foundation model training. By introducing a stable data distillation method for medical image analysis through a progressive trajectory matching strategy, this approach offers a privacy-preserving data-sharing mechanism for pretraining foundation models \cite{yu2024progressive}. Another solution for privacy leakage caused by data sharing is to use synthesized images. The utilization of artificially synthesized images along with corresponding textual data for training enables the medical foundation model to effectively absorb knowledge of disease symptomatology, thereby mitigating potential breaches of patient confidentiality \cite{wei2024visionclip}. To satisfy ``the right to be forgotten'' in the regulations \cite{goddard2017eu}, seamless data modification is utilized to improve privacy in medical image analysis without model retraining, thus reducing the risk of privacy breaches \cite{doerrich2024integrating}.\\

\subsubsection{Robustness in Disease Diagnosis}
\noindent\textbf{Vulnerabilities  to adversarial attacks, backdoor attacks, and weight manipulations.} Although foundation models for medical image analysis achieved great success, the robustness of foundation models for medical image analysis is still considerable. Veerla~\etal\cite{veerla2024vulnerabilities} explores the vulnerabilities of the Pathology Language-Image Pretraining (PLIP) model by employing Projected Gradient Descent (PGD) adversarial attacks to intentionally induce misclassifications. The findings of the study emphasize the pressing need for robust defenses to ensure the security of foundation models for medical image analysis. Jin~\etal\cite{jin2024backdoor} point out that backdoor attacks pose a threat to MedCLIP and identify vulnerabilities like BadMatch, which exploits minor label discrepancies. The study reveals that current defenses are inadequate against such backdoor attacks in medical foundation models. 
Another study also finds out that manipulating just 1.1\% of an LLM's weights allows for the injection of inaccurate biomedical facts that propagate throughout its output without impacting its performance on other tasks \cite{han2023medical}. This susceptibility highlights serious security and trustworthiness concerns regarding the utilization of LLMs in healthcare.\\

\subsubsection{Reliability in Disease Diagnosis}
\noindent\textbf{Satety evaluation of applying GPT-4 for medical domain.} Similarly, ensuring the reliability of foundation models in medical image analysis is crucial, necessitating measures to prevent hallucination and the propagation of outdated information.
Wu~\etal \cite{wu2023can} assess the performance of GPT-4 by OpenAI in multimodal medical diagnosis, they found that GPT-4 shows its safety guarantees against potential misuse and ensures users are aware of its capabilities and limitations. Specifically, when GPT-4 is asked to make a diagnosis, such as providing a diagnosis for a chest X-ray, it incorporates safeguards: refusal to offer a diagnosis, emphasis on limitations, and expression of uncertainty.

\subsubsection{Explainability in Disease Diagnosis}
\noindent\textbf{Enhancing explainability by concept-bottleneck models.} As a high-stakes domain with rigorous demands of trustworthiness, foundation models used for disease diagnosis must not only demonstrate high performance but also fulfill the criteria for explainability. 
To meet the explainable requirements and inspired by description-based interpretable approach \cite{menon2022visual}, one recent study proposes a framework for explainable zero-shot medical image classification utilizing vision-language models like CLIP along with LLMs like ChatGPT \cite{liu2023chatgpt}. The primary concept involves harnessing ChatGPT's capabilities to automatically generate comprehensive textual descriptions encompassing disease symptoms and visual attributes, moving beyond mere disease labels. This supplementary textual data enhances the precision and interpretability of diagnoses generated by CLIP.
By aligning the semantics of images, learnable prompts, and clinical concept-driven prompts at various levels of detail, Bie~\etal\cite{bie2024xcoop} propose a novel explainable prompt learning framework, leveraging medical expertise. Agarwal~\etal\cite{agarwal2023representing} presents a novel explainability strategy in healthcare, leveraging a vision-language model to identify language-based descriptors of visual classification tasks, demonstrating alignment with clinical knowledge and potential for aiding non-expert human understanding of specialized medical tasks. In pursuit of explainability in deep learning, Doerrich~\etal\cite{doerrich2024integrating} integrate a k-Nearest Neighbor (k-NN) classifier with a vision-based foundation model, enhancing both interpretability and adaptability.
Yan~\etal\cite{yan2023robust} propose a new paradigm to build interpretable medical image classifiers with natural language concepts, wherein they query clinical concepts from GPT-4 and transform latent image features into explicit concepts using a vision-language model.
Wang \etal \cite{wang2024copilotcadempoweringradiologistsreport} introduce CopilotCAD, which combines the foundation model's computational power with the expertise of radiologists, offering a user-friendly interface for interactive, image-based diagnostics. It empowers radiologists to make informed decisions with the support of AI-generated quantitative data and visual aids, enhancing the explainability and transparency of CAD systems.

\subsubsection{Fairness in Disease Diagnosis}
\noindent\textbf{Disparate model performance on marginalized groups.} In the realm of disease diagnosis, ensuring fairness is significant as we delve into the intricate interplay between medical algorithms and demographic attributes. Some experimental results demonstrate that some foundation models for healthcare are considered unfair to some extent. 
Chen~\etal\cite{chen2024chexagent} examine their proposed foundation model,~\emph{CheXagent}, for Chest X-ray interpretation, conducting a fairness evaluation, in which they examine different groups of people according to their sex, race and age. The results unveil disparities in model performance across various groups and underscore the presence of biases. Furthermore, Yang~\etal\cite{yang2024demographic} investigate the fairness of the existing Vision-language (VL) foundation models in chest X-ray diagnosis across five globally-sourced datasets. Their findings unveil a consistent pattern: these VL models consistently underdiagnose marginalized groups, with even higher rates observed in intersectional subgroups, such as Black female patients, when compared to diagnoses by board-certified radiologists.\\

\noindent\textbf{New dataset for fairness evaluation fairness} Inspired by some widely-used VL models such as CLIP, VL models for medical tasks such as MedCLIP also have attracted a lot of attention \cite{zhao2023clip}. However, the absence of medical VL datasets poses a significant challenge for studying fairness. Luo~\etal\cite{luo2024fairclip} addressed this issue by introducing \emph{Harvard-FairVLMed} dataset. This dataset offers comprehensive demographic attributes, ground-truth labels, and clinical notes, enabling a thorough investigation into fairness within VL foundational models. 
Based on \emph{Harvard-FairVLMed} dataset, some fairness issues are found when using two widely-used VL models (CLIP and BLIP2).\\

\noindent\textbf{To mitigate fairness issues, methods such as FairCLIP and Universal Debiased Editing are proposed.} To improve the fairness of VL models, an optimal-transport-based method named \emph{FairCLIP} is proposed. It aims to trade off balance between model performance and fairness by minimizing the distribution gaps across both the overall dataset and individual demographic groups. Considering the scenario of using foundation model API, in which we have very limited model control and computational resources, Jin~\etal\cite{jin2024universal} propose a Universal Debiased Editing (UDE) strategy. This method is applicable to both white-box and black-box foundation model APIs, with the capability to mitigate bias within both the foundation model API embedding and the images themselves. In addition, the fairness in privacy-preserving foundation models for medical images is investigated. Berrada~\etal\cite{berrada2023unlocking} examine pre-trained foundation models fine-tuned with Differential Privacy (DP) using two medical imaging benchmarks. Evaluation results show that private medical classifiers do not exhibit larger performance disparities across demographic groups than non-private models, making DP training a practical and reliable approach.

\section{Challenges and Future Directions} 
\label{challenges}
In the field of medical image analysis, the integration of foundation models heralds transformative potential, yet it is fraught with a complex tapestry of challenges that underscore the paramount importance of trustworthiness. As we venture into this new era, the intricacies of ensuring the trustworthiness of AI systems become increasingly salient. This section aims to dissect multifaceted challenges and possible future directions that promise to navigate these hurdles, thereby paving the way for the responsible and effective deployment of foundation models in the realm of medical image analysis.

\subsection{Datasets and Benchmarks}
Ensuring the trustworthiness of foundation models in medical image analysis is challenging due to their complexity and the critical nature of healthcare decisions, which usually need human inputs. The variability of human inputs and the subjective nature of medical data labeling further complicate the validation of these foundation models, as it can lead to inconsistent ground truths. Ensuring that the model generalizes well across diverse patient populations and imaging techniques without a comprehensive way to test all possible scenarios adds to the verification challenge. In addressing the trustworthiness of foundation models in medical image analysis, the development of high-quality datasets and benchmarks is paramount. Ground truth data serves as a cornerstone not only for model training but also for the evaluation of models in terms of safety, robustness, fairness, and reliability. Achieving ground truth in medical datasets often requires a multidisciplinary approach that includes consensus from various medical experts, rigorous patient outcome tracking, and potentially the use of synthetic data where real-world data is scarce or ethically challenging to obtain. Benchmarks need to extend beyond mere diagnostic accuracy and must encompass clinical relevance, ensuring models are not only technically sound but also deliver real-world benefits to patient care. They must also account for fairness, to avoid perpetuating existing biases, and robustness, to ensure models remain reliable under diverse and unpredictable clinical conditions. This comprehensive approach to datasets and benchmarks will foster the development of models that are truly trustworthy, capable of integrating seamlessly into clinical workflows and contributing positively to patient outcomes.

Currently, FairMedFM~\cite{jin2024fairmedfm} is proposed to address the lack of comprehensive benchmarks by offering an integrated framework that evaluates fairness in medical imaging foundation models. By integrating 17 popular medical imaging datasets and analyzing 20 widely used foundation models, it reveals the existence of bias, fairness-utility tradeoffs, inherent fairness issues in the datasets, and limited effectiveness of existing mitigation methods.

\subsection{Hallucinations}
One of the emerging challenges in foundation models for medical imaging is the issue of hallucinations, where the model erroneously identifies or interprets features in the medical images that are not present or generates plausible but incorrect information. This can lead to inaccurate diagnoses and potentially harmful recommendations. Addressing the issue of hallucinations in foundation models used for medical image analysis is critical for ensuring their reliability and trustworthiness in healthcare settings. It necessitates enhancing the models' training with more diverse and comprehensive datasets that better capture the variability in real-world medical scenarios. It also requires the development of advanced interpretability techniques to understand the decision-making processes of these models. Rigorous validation frameworks and continuous feedback mechanisms involving expert clinicians are essential to identify, mitigate, and correct hallucinations, thereby improving the models' diagnostic accuracy and ensuring safer deployment in clinical environments.


\subsection{Alignment}
A critical future direction for foundation models in medical imaging is enhancing their alignment with clinical workflows and ethical standards. Ensuring that these models produce outputs that are not only accurate but also clinically relevant is essential. This involves refining the models to understand and adhere to medical protocols and integrating continuous feedback from medical experts. To achieve this, integrating reinforcement learning with human feedback (RLHF) into medical image analysis is a promising approach. However, unlike general applications, integrating RLHF into medical image analysis for alignment faces several challenges~\cite{casper2023open,kaufmann2023survey}.
 Firstly, the high annotation costs associated with acquiring expert feedback from medical professionals, such as doctors or radiologists, pose a significant barrier. Their specialized knowledge is essential for effective tuning of the model, but their time and effort are limited resources, making it difficult to obtain sufficient annotations. Secondly, there are risks of introducing new biases into the model when relying on feedback from a limited number of experts. These biases may stem from subjective and inconsistent diagnoses, potentially complicating the objective assessment of medical images. Additionally, privacy concerns surrounding medical images, which often contain sensitive personal health information, further complicate the acquisition of well-annotated data for RLHF. Legal and ethical challenges may arise, particularly when attempting to obtain patient feedback or expert judgments for training purposes.

\subsection{Regulatory and Ethical Frameworks}
Establishing robust regulatory and ethical frameworks for foundation models in healthcare is essential to navigate the complex interplay between advancing technology and patient welfare. These frameworks must ensure that patient consent is informed and respected, safeguarding data privacy in accordance with stringent standards. Transparency in model operations must be mandated to allow for interpretability and justifiable reliance on AI-driven decisions. Accountability measures need to be clear and enforceable, delineating responsibilities in the event of diagnostic errors or mishandling of patient data. Furthermore, these guidelines should promote fairness and prevent the widening of healthcare disparities, making certain that the benefits of AI are accessible across diverse patient groups. Engaging a broad coalition of stakeholders, including healthcare professionals, legal experts, AI developers, and patient rights advocates, is crucial in formulating these frameworks to ensure they are ethically sound, socially responsible, and adaptable to the rapid evolution of AI in medicine.

\subsection{Interdisciplinary Collaboration}
Incorporating interdisciplinary knowledge into foundation models presents a significant challenge yet is crucial for enhancing their trustworthiness in medical image analysis. The necessity to synthesize diverse perspectives—from computer science, ethics, medicine, to law—into a cohesive learning framework for these models underlines a complex integration challenge. Each discipline brings its own methodologies, terminologies, and priorities, making the harmonization of this knowledge within the AI development process non-trivial. Additionally, the dynamic nature of medical knowledge, ethical standards, and regulatory requirements further complicates this task, requiring models to continuously adapt to new information and changing guidelines. This challenge not only demands advanced technical solutions for integrating multifaceted inputs but also calls for new collaborative structures that facilitate effective communication and knowledge exchange among experts from various fields. Overcoming these hurdles is essential for building foundation models that are not only technically proficient but also ethically responsible, clinically relevant, and legally compliant, thereby truly earning the trust of users and stakeholders in healthcare settings.

\section{Conclusion}
\label{conclusion}
In conclusion, this work presents the first comprehensive analysis of foundation models in medical image analysis, focusing on the potential trustworthiness issues essential for responsible healthcare applications. Foundation models show transformative potential, enhancing tasks like segmentation, report generation, medical Q\&A, and disease diagnosis. However, our survey highlights significant concerns regarding their trustworthiness, particularly in privacy, robustness, reliability, explainability, and fairness.

We categorize and evaluate trustworthiness across these dimensions, offering a nuanced view of where foundation models excel and where limitations persist. This analysis contributes a foundational map of the current state of foundation models in medical image analysis, outlining critical challenges and actionable directions for future research. Specifically, addressing these trustworthiness challenges is paramount to harnessing foundation models' full potential to improve healthcare outcomes.

The path forward requires efforts on building standardized datasets and benchmark for evaluation, innovations that align technical progress with ethical responsibility and equity, ensuring that foundation models in healthcare not only advance performance but also adhere to principles that safeguard patient trust and societal impact.

\bibliographystyle{ieeetr}
\bibliography{references.bib}

\end{document}